\documentclass{elsart}

\usepackage{natbib}
\usepackage{amsmath}
\usepackage{amssymb}
\usepackage{graphicx}
\usepackage{rotating}

\journal{}

\begin{document}

\begin{frontmatter}

\title{Face Recognition using Principal Component Analysis and Log-Gabor Filters}

\author{Vytautas Perlibakas}

\ead{vperlib@mmlab.ktu.lt}

\address{
Image Processing and Analysis Laboratory, Computational Technologies Centre, Kaunas University of Technology,
Studentu st. 56-305, LT-51424 Kaunas, Lithuania
}

\begin{abstract}

In this article we propose a novel face recognition method based
on Principal Component Analysis (PCA) and Log-Gabor filters.
The main advantages of the proposed method are its simple implementation, training, and very high recognition accuracy.
For recognition experiments we used 5151 face images of 1311 persons from different sets of the
FERET and AR databases that allow to analyze how recognition accuracy is affected by the change of facial expressions,
illumination, and aging.
Recognition experiments with the FERET database (containing photographs of 1196 persons)
showed that our method can achieve maximal 97-98\% first one recognition rate and 0.3-0.4\% Equal Error Rate.
The experiments also showed that the accuracy of our method is less affected by eye location errors
and used image normalization method than of traditional PCA -based recognition method.

\end{abstract}


\end{frontmatter}

\section{Introduction}

Principal Component Analysis (PCA) or Karhunen Loeve Transform (KLT) - based face recognition
method was proposed in ~\citep{Turk1991} and became very popular because of its relatively simple implementation
and high recognition accuracy. During past fifteen years face recognition was a field of active research, and
many other statistical methods (related to PCA) were investigated and proposed for face recognition:
Linear Discriminant Analysis (LDA), Independent Component Analysis (ICA) ~\citep{Bartlett2002},
Kernel PCA, Dual PCA ~\citep{Moghaddam2002}.
Because KLT is data dependent and is not very fast, other transforms were also used for face recognition:
Discrete Cosine Transform (DCT) ~\citep{Hafed2001}, Fast Fourier Transform (FFT) ~\citep{Spies2000},
Discrete Wavelet Transform (DWT) ~\citep{Feng2000}, Wavelet Packet Decomposition (WPD) ~\citep{Garcia2000}.
~\citet{Wiskott1997} proposed Elastic Bunch Graph Matching (EBGM) and Gabor wavelets -based face recognition method that achieved
very high recognition accuracy. ~\citet{Escobar2002} used EBGM -based recognition of faces in Log-Polar coordinates.
Comprehensive overview of various face recognition methods could be found in ~\citep{Zhao2000}.
As it was shown by numerous experiments, face recognition accuracy can be increased by combining several
methods, for example, DCT+PCA ~\citep{Ramasubramanian2001}, DWT+PCA ~\citep{Feng2000}, WPD+PCA ~\citep{Perlibakas2004},
PCA+LDA ~\citep{Zhao1998}, or by using various image pre-processing methods.
Recent results also showed that using Gabor or Log-Gabor features instead of traditional
greyscale features and by combining these features with well known recognition methods
like PCA, ICA, LDA or SVM it is possible to achieve very high recognition acuracy.
Now we will overview various combined face recognition methods that
use Gabor or Log-Gabor features and that are related with a method that we propose
in this publication. Although many researchers used the same Gabor filters and well known
feature compression and classification methods, all proposed methods differ from each other
by feature selection techniques, parameters of filters, used classification method and its parameters,
distance measure, and image normalization method.
After filtering face image with Gabor filters of multiple scales (usually 4-5) and
orientations (usually 6-8) we get very large number of features (24-40 images of the same
dimensions as initial image). Perhaps one of the most important
questions is how to reduce the number of these features for further processing.
The most popular method is to extract Gabor features at a small number (usually less than 100)
of face points around face features (like eyes, lips, nose) that were detected
using EBGM ~\citep{Wiskott1997} or similar method.
At each detected point are extracted Gabor features from all scales and orientations.
~\citet{Lyons2000} combined EBGM Gabor features and LDA-based recognition method.
~\citet{Smeraldi2002} detected face features using saccadic search with a set of
Log-Gabor filters that were arranged to concentric circles (retinas). At detected
points were extracted Log-Gabor features and passed to the SVM -based classifier.
EBGM -based feature selection was also used by ~\citet{Wang2003} for Bayesian PCA -based
recognition.
Because EBGM -based methods require training with manually labelled faces, other researchers use
feature extraction methods that do not require such training.
One possible approach is to combine Gabor magnitude images from all scales and
orientations to a single feature vector (image) and use this vector for recognition. Such method
was used by ~\citet{Zhang2004} for Gabor+AdaBoost face recognition.
Because such combined feature vectors may be too large for further processing, we can reduce
the number of features by using smaller initial images.
~\citet{Liu2002} decided to downsample feature images of each scale and resolution and then
combine these features to a single vector for Gabor+Enhanced LDA ~\citep{Liu2002} and
Gabor+ICA ~\citep{Liu2003} -based recognition using $L_{1}$, Euclidean, and cosine -based distance measures.
In order to reduce the number of Gabor features, ~\citet{Kepenekci2002} used sliding window based
search at all scales and orientations. In each window were extracted features with maximal
magnitudes, stored their locations, and then both magnitudes and locations were used for
comparison. For recognition was used very similar distance measure that was used by
~\citep{Wiskott1997} with additional constraints to feature locations.

In this article we propose to find the locations of Log-Gabor ~\citep{Field1987} features
with maximal magnitudes at single scale and multiple orientations using sliding window
-based search and then use the same feature locations for all other scales. 
For further feature compression we used Principal Component Analysis (PCA) because its
simple implementation, fast training and because using PCA with
"whitened" angle -based distance measure it is possible to achieve similar recognition accuracy 
like using EBGM and LDA -based recognition methods ~\citep{CSU2003}.
We tested our method using 5151 face images of 1311 persons from
the FERET and AR databases and the results showed that the proposed recognition method
can achieve higher recognition accuracy than many other existing methods.
The results of experiments also showed that PCA using Log-Gabor features is less sensitive to
face detection errors and used image normalization method than PCA using greyscale features.

\section{Feature extraction using Log-Gabor filters}

The Log-Gabor filters were proposed by ~\citet{Field1987} for coding of natural images.
The experiments showed, that these filters are consistent with the measurements
of the mammalian visual system and are more suitable for coding of natural images than
~\citet{Gabor1946} filters.
The Log-Gabor filter in frequency domain can be constructed in terms of
two components, namely the radial filter component $G(f)$ and the angular filter component $G(\theta )$.
In polar coordinates the filter transfer function could be written in the following form
~\citep{Field1987}, ~\citep{Bigun1994}, ~\citep{Kovesi1996}:
\begin{equation}
G(f,\theta ) = G(f) \cdot G(\theta ) = \exp \left( { - \frac{{(\log ({f \mathord{\left/
 {\vphantom {f {f_0 }}} \right.
 \kern-\nulldelimiterspace} {f_0 }}))^2 }}{{2(\log ({k \mathord{\left/
 {\vphantom {k {f_0 }}} \right.
 \kern-\nulldelimiterspace} {f_0 }}))^2 }}} \right) \cdot \exp \left( { - \frac{{(\theta  - \theta _o )^2 }}{{2\sigma _\theta ^2 }}} \right), \label{eq_filter}
\end{equation}
here
$f_0$ is the centre frequency of the filter, $k$ determines the bandwidth of the filter, 
$\theta _o$ is the orientation angle of the filter, and
$\sigma _\theta   = {{\vartriangle \theta } \mathord{\left/
 {\vphantom {{\vartriangle \theta } {s_\theta  }}} \right.
 \kern-\nulldelimiterspace} {s_\theta  }}$
where $s_\theta$ - scaling factor, $\vartriangle \theta$ - orientation spacing between filters.
For face recognition we generated multiple Log-Gabor filters of different scales and orientations using the following
parameters:
$f_0  = {1 \mathord{\left/
 {\vphantom {1 \lambda }} \right.
 \kern-\nulldelimiterspace} \lambda }$,
$\lambda  = \lambda _0  \cdot s_\lambda ^{(n_s  - 1)} $,
${k \mathord{\left/
 {\vphantom {k {f_0 }}} \right.
 \kern-\nulldelimiterspace} {f_0 }} = \sigma _f $,
$n_s  = 1,...,N_s$;
$\theta _o  = {{\pi (n_o  - 1)} \mathord{\left/
 {\vphantom {{\pi (n_o  - 1)} {N_o }}} \right.
 \kern-\nulldelimiterspace} {N_o }}$,
$\vartriangle \theta  = {\pi  \mathord{\left/
 {\vphantom {\pi  {N_o }}} \right.
 \kern-\nulldelimiterspace} {N_o }}$,
$n_o = 1,...,N_o$;
$\lambda _0 = 5$,
$s_\lambda = 1.6$,
$\sigma _f = 0.75$,
$N_s  = 4$,
$s_\theta = 1.5$,
$N_o = 6$,
here 
$\lambda _0$ is the wavelength of the smallest scale filter,
$s_\lambda$ is the scaling factor between successive filter scales,
$N_s$ is the number of scales,
$N_o$ is the number of orientations.
Most of these parameters were chosen following the recommendations of ~\citep{Kovesi2003}.

Using Eq. ~\ref{eq_filter} we calculate two-dimensional Log-Gabor filter $G_{n_{o}, n_{s}}$ in Fourier space
of a chosen filter scale and orientation. The size of the filter array $G_{n_{o}, n_{s}}$ is the same as the size of
the two-dimensional image $I$ that we wish to filter. Then we perform filtering (convolution in Fourier space), magnitude calculation and
masking using the following equation:
\begin{equation}
V_{n_{o}, n_{s}}=abs(IFFT2(G_{n_{o}, n_{s}}.*FFT2(I))).*mask,
\end{equation}
here
"$.*$" - array (not matrix) multiplication,
$I$ - normalized (cropped, masked) face image,
$G_{n_{o}, n_{s}}$ - Log-Gabor filter of desired orientation and scale in Fourier space,
$FFT2$ - two-dimensional Fast Fourier Transform, $IFFT2$ - inverse $FFT2$,
$mask$ - binary mask for masking magnitude image (the same as is used for masking greyscale face image $I$ in order
to leave only the internal part of the face), $V_{n_{o}, n_{s}}$ - masked Log-Gabor magnitude image.

\begin{figure}[htbp]
\centering \includegraphics[width=120mm]{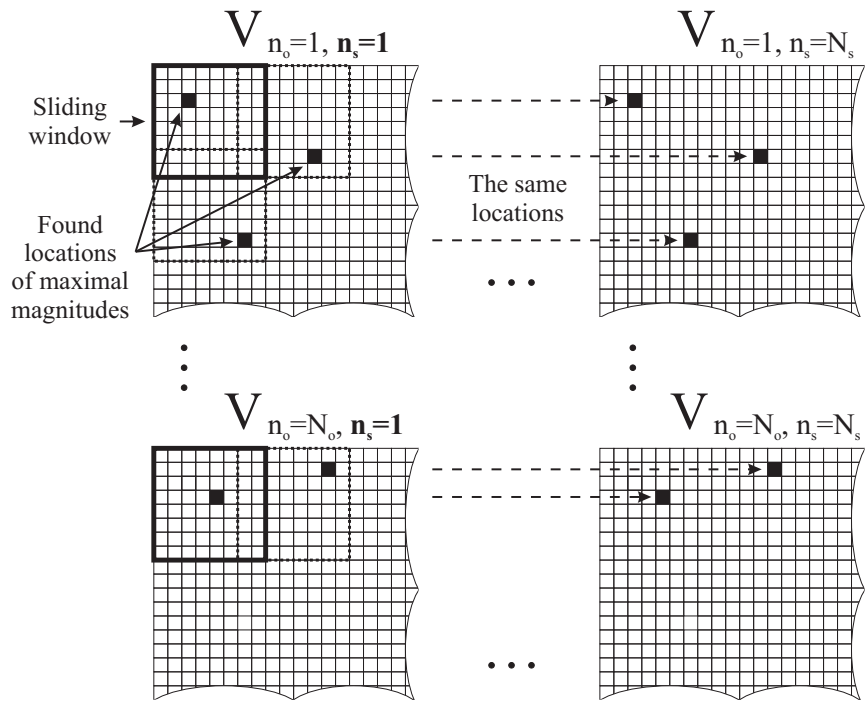}
\caption{Selection of Log-Gabor magnitude features using sliding window algorithm}
\label{fig_sliding_window}
\end{figure}

After image filtering with multiple Log-Gabor filters ($N_{s}$ scales and $N_{o}$ orientations) we get very
large number of Log-Gabor features (magnitude values in all $N_{s} \cdot N_{o}$ magnitude images).
In order to reduce the number of features and achieve partial face recognition invariance
with respect to different facial expressions and minor face detection errors,
we use sliding window algorithm that is illustrated in Fig. ~\ref{fig_sliding_window}.
Rectangular window of a chosen size (e.g., 8x8 pixels) is slided over the magnitude image $V_{n_{o}, 1}$ using 
some sliding step (e.g., 6 pixels, overlapping of windows is 8-6=2 pixels). In each window we find one maximal magnitude value and remember
the location (coordinates in image $V_{n_{o}, 1}$) of this value. If several equal values are found, we use the one that
is closer to the centre of the window. If magnitude image is masked, we perform search only in an unmasked image
part.

We apply sliding window algorithm only for the first scale ($n_{s}=1$)
(obtained using filter with the smallest chosen wavelength) of each orientation
and find the locations of highest magnitudes. Then these locations are used for extracting magnitudes from other scales,
corresponding to the analysed orientation, as it is illustrated in Fig. ~\ref{fig_sliding_window}.
In this figure feature locations are marked with black points.
Using scales $n_{s}=1$ we decide at what locations (coordinates) we will extract the features
and then use the same coordinates for all magnitude images, corresponding
to the same processed orientation $n_{o}$ (the same orientations $n_{o}$ - the same locations,
different orientations - different locations). All extracted Log-Gabor features (magnitude values)
are stored in a one-dimensional vector $X$ and used as input for Principal Component Analysis -based face recognition method.

\begin{figure}[htbp]
\centering \includegraphics[width=110mm]{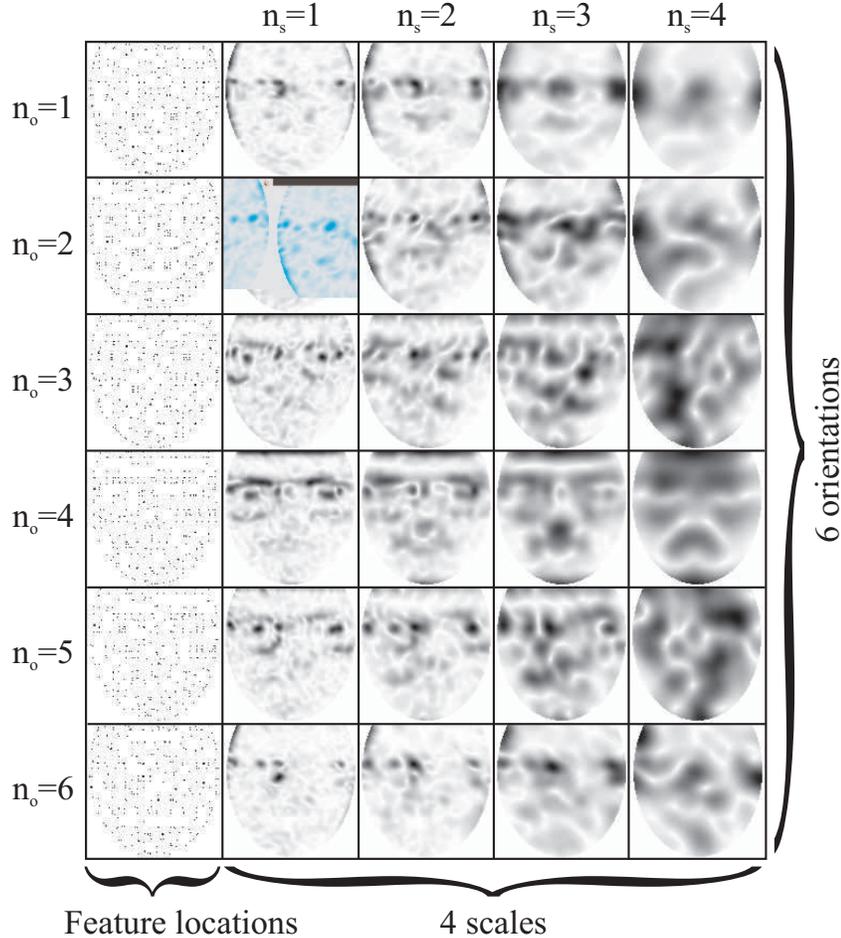}
\caption{Image magnitudes after using Log-Gabor filters, and the locations of features at different orientations that were found using sliding window algorithm}
\label{fig_magnitudes}
\end{figure}

Example Log-Gabor magnitude images are presented in Fig. ~\ref{fig_magnitudes} (images are inverted, dark points mean high magnitude values).
In this example we used filters of $N_{o}=6$ orientations and $N_{s}=4$ scales ($N_{o} \cdot N_{s} = 24$ filters), and
filtered a normalized (derotated, masked) facial image with these filters.
Calculated Log-Gabor magnitude images were also masked. Left-most binary images in 
Fig. ~\ref{fig_magnitudes} show the locations (black points) of Log-Gabor features that were found
using sliding window algorithm.
It must be noted, that for the selected image size the Log-Gabor filters
(of different sizes and orientations) can be calculated only once and stored.
When we perform face recognition, the Log-Gabor features (found using sliding window)
for each image from the database of faces are also calculated only once and stored.

\section{Face recognition using Principal Component Analysis of Log-Gabor features}

In this section we will describe Karhunen-Loeve transform (KLT)
-based face recognition method, that is often called Principal Component Analysis (PCA).
We will present only the main formulas of this method, which details could be
found in ~\citep{Gross1994}.

Let $X_j$  be $N$-element one-dimensional image-column (vector) and suppose that we have $r$
such images ($j = 1,...,r$).
In traditional PCA -based face recognition method, these images contain grey values of the two-dimensional
facial photographs. \emph{In our case these one-dimensional images $X_j$ (data vectors) contain Log-Gabor features}.
We calculate the mean vector, centred data vectors and covariance matrix:
$m = \frac{1}{r}\sum\limits_{j = 1}^r {X_j},$
$d_j = X_j - m,$
$C = \frac{1}{r}\sum\limits_{j = 1}^r d_j d_j^T,$
here
$X = (x_1 ,x_2 ,...,x_N )^T,$
$m = (m_1 ,m_2 ,...,m_N )^T,$
$d = (d_1 ,d_2 ,...,d_N )^T.$

In order to perform KLT, it is necessary to find eigenvectors $u_k$ and eigenvalues $\lambda _k$
of the covariance matrix $C$ ( $Cu_k  = \lambda _k u_k$ ). Because
the dimensionality ($N^2$) of the matrix $C$ is usually large even for small
images, and computation of the eigenvectors using traditional methods is complicated, dimensionality
of matrix $C$ is reduced using the decomposition described in ~\citep{Kirby1990} (if the
number of training images is smaller than the length of the vector $X$).
Found eigenvectors $u = (u_1 ,u_2 ,...,u_N )^T$ are normed and sorted in decreasing order
according to the corresponding eigenvalues. Then these vectors are transposed and arranged to form the row-vectors of the
transformation matrix $T$. Now any data $X$ can be projected into the eigenspace and
"whitened" ~\citep{Bishop1995} using the following formula:
\begin{equation}
Y = \Lambda ^{ - {1 \mathord{\left/
 {\vphantom {1 2}} \right.
 \kern-\nulldelimiterspace} 2}} T(X - m),
\end{equation}
here
$X = (x_1 ,x_2 ,...,x_N )^T,$
$Y = (y_1 ,y_2 ,...,y_r ,0,...,0)^T,$

$
\Lambda ^{ - {1 \mathord{\left/
 {\vphantom {1 2}} \right.
 \kern-\nulldelimiterspace} 2}}  = diag(\sqrt {{1 \mathord{\left/
 {\vphantom {1 {\lambda _1 }}} \right.
 \kern-\nulldelimiterspace} {\lambda _1 }}} ,\sqrt {{1 \mathord{\left/
 {\vphantom {1 {\lambda _2 }}} \right.
 \kern-\nulldelimiterspace} {\lambda _2 }}} ,...,\sqrt {{1 \mathord{\left/
 {\vphantom {1 {\lambda _r }}} \right.
 \kern-\nulldelimiterspace} {\lambda _r }}} ).
$

For projection we can use not all found eigenvectors, but only a few of them,
corresponding to the largest eigenvalues. We can manually select the desired number of eigenvectors
or use the method described in ~\citep{Swets1998}.

For each facial image (that we wish to use for recognition) we find Log-Gabor features, projected these features
into the eigenspace and calculate eigenfeature
vector $Z = (z_1 ,z_2 ,...,z_n )^T  = (y_1 ,y_2 ,...,y_n )^T$, here $n$ is the number
of features. Recognition of unknown face is performed by 
calculating its feature vector $Z_{new}$ and comparing it with the feature vectors of known faces. 
For comparison we calculate the distances $\varepsilon _i ( {Z}_{{\rm{new}}}, {Z}_i )$ between
unknown face and each known face 
and say that the face with feature vector ${Z}_{new}$ belongs to a person $s = \arg \mathop {\min }\limits_i \left[ {\varepsilon _i } \right]$.
For rejection of unknown faces a threshold ${\rm{\tau }}$ is chosen and it is said that the
face with projection ${Z}_{new}$ is unknown if $\varepsilon _s  \ge {\rm{\tau }}$.
For recognition we used cosine-based distance measure $\varepsilon _i ( {Z}_{{\rm{new}}}, {Z}_i ) = - cos ( {Z}_{{\rm{new}}}, {Z}_i )$,
because using this distance measure we can achieve higher recognition accuracy ~\citep{Perlibakas2004}
than using the Euclidean or Manhattan distance measures.

\section{Normalization of face images}

\begin{figure}[h]
\centering \includegraphics[width=110mm]{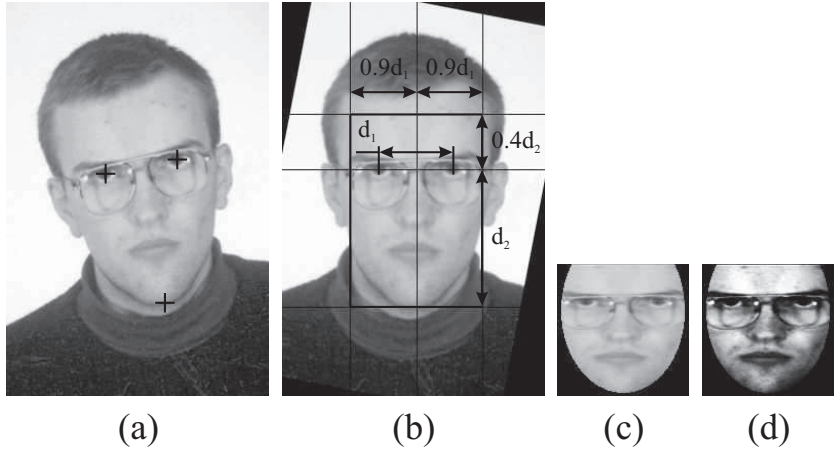}
\caption{Image normalization (a-d - 3-point method, e - 2-point method): a) initial image with selected eyes and chin b) denoised, derotated image, and cropping schema c) cropped, resized and masked image d) normalized image after histogram equalization}
\label{fig_image_normalization}
\end{figure}

For recognition experiments we used two image normalization methods.
One method uses manually selected centres of eyes and the tip of chin (3-point normalization method),
and another method for normalization uses only the centres of eyes (2-point normalization method).
Image normalization procedure of 3-point method is presented in Fig. ~\ref{fig_image_normalization}.
The last image (Fig. ~\ref{fig_image_normalization} (e)) also presents the result of 2-point normalization. 
For illustration we used an image from our personal archive.

Now we will describe our implementation of 3-point normalization method.
Initial images were denoised (using Gaussian filter with $\sigma=0.5$ and window size 5x5),
derotated (in order to make the line connecting eye centres horizontal), cropped, resized (to the size of 128x128 pixels),
masked.
For rotation and resizing we used bicubic interpolation.
For masking we used an ellipse with 
central point (64.5,45.5), horizontal axis of 120 pixels, and vertical axis of 160 pixels.
Then for unmasked part of the image we performed histogram equalization (256 levels).
When the image is masked, are left 12646 unmasked pixels of 16384 (128x128).

For initial comparison of PCA and Log-Gabor PCA methods we used 3-point normalization method
in order to perform experiments with faces that are not overcropped and also contain no scene's background
information. 
Because the tip of chin may be hard to locate, most recognition methods for normalization use
only the centres of eyes.
So we used 3-point normalization only for initial comparison of PCA and Log-Gabor PCA methods, 
and for the rest of experiments we used 2-point normalization method (for normalization are used only the centres of eyes).
Because there is no agreement how images should be normalized for face recognition experiments,
we implemented 2-point normalization that is very similar to the ~\citep{CSU2003} normalization method.
Similar method was also used by some participants of the FERET ~\citep{NIST2001} tests.
At first images are derotated in order to make the line connecting eye centres horizontal.
Then images are resized in order to make the distance between eyes equal to 70 pixels and
cropped to the size of 130x150 pixels. During cropping the centres of eyes are vertically
positioned on y=45 line (the centre of coordinates (0,0) is in the left top corner).
Then the image is masked using 
an ellipse with its central point (65.5,50.5), horizontal axis of 128 pixels,
and vertical axis of 236 pixels.
After masking are left 17237 unmasked pixels.
For an unmasked image part is performed histogram equalization.
The main differences between our 2-point normalization and ~\citep{CSU2003} normalization are as follows:
initial images we filtered using Gaussian filter (CSU used no filtering),
for image rotation and resizing we used bicubic interpolation (CSU used bilinear interpolation),
cropped 130x150 images we resized to 128x128 pixels and then masked with resized (to 128x128)
binary mask.
After such masking were left 14454 unmasked pixels. Then for an unmasked image part we
performed histogram equalization. The result of this normalization is presented in
Fig. ~\ref{fig_image_normalization} (e).

When we performed face recognition using Log-Gabor PCA method, masked face images were filtered
with Log-Gabor filters (24 filters of 6 orientations and 4 scales) and calculated magnitude images.
We masked these Log-Gabor magnitude images using the same masks that were used
for image normalization and performed sliding window search of Log-Gabor features.
Search window size is 8x8 pixels, sliding step is 6 pixels (the same in horizontal and vertical
directions), and window overlap is 8-6=2 pixels.
After using sliding window algorithm with masked magnitude
images, we select 9240 magnitude values (Log-Gabor features) for 3-point normalization method and
10008 magnitude values for 2-point normalization method, that
are located in the unmasked parts of magnitude images. This is the total number of values in all
6 orientations and 4 scales.
Also we can use an unmasked magnitude images (initial greyscale images are always masked), perform
sliding window search in a whole magnitude image and select in total 10584 features
for both 3-point and 2-point methods (the size of images is the same).
In all the experiments we used the same normalized image patterns and the same implementation of PCA.
The distances between "whitened" feature vectors were measured using cosine -based distance measure.

\section{Used recognition performance measures}

\begin{figure}[ht]
\begin{minipage}[t]{0.5\linewidth}
\centering \includegraphics[height=76mm]{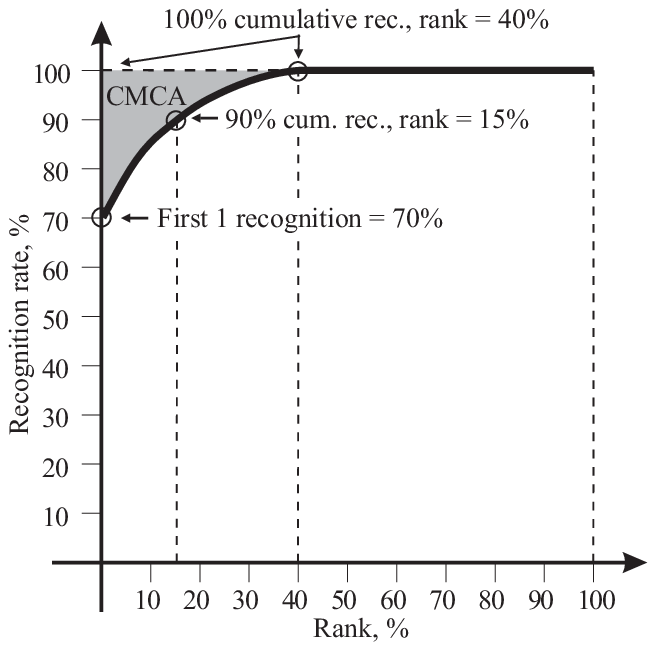}
\caption{Cumulative Match Characteristic (CMC)}
\label{fig_cmc}
\end{minipage}%
\hspace{5mm}
\begin{minipage}[t]{0.5\linewidth}
\centering \includegraphics[height=76mm]{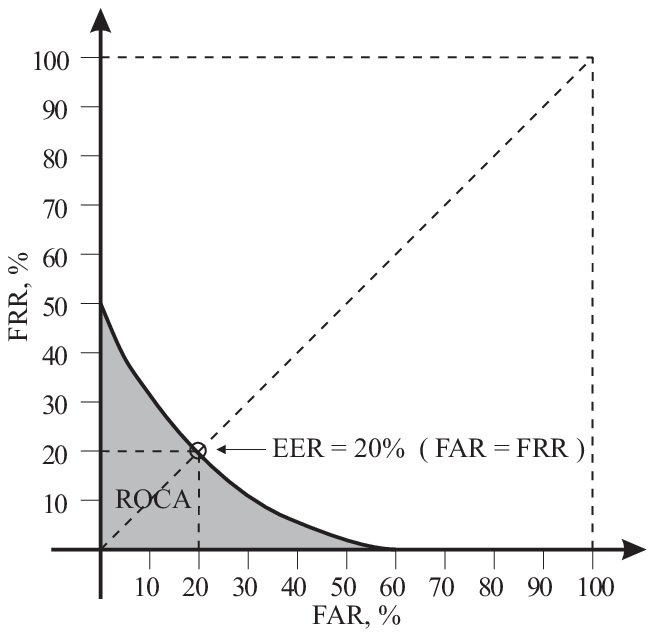}
\caption{Receiver Operating Characteristic (ROC)}
\label{fig_roc}
\end{minipage}
\end{figure}

For comparison of face recognition methods we used
Cumulative Match Characteristic (CMC) and Receiver Operating
Characteristic (ROC) - based measures described in ~\citep{Bromba2003}:
the area above Cumulative Match Characteristic (CMCA) (smaller CMCA means better overall recognition accuracy);
how many images (in percents) must be extracted from the database in order to achieve some cumulative
recognition rate (e.g., not smaller than 95-100\%) (smaller values mean that we need to extract fewer images in
order to achieve some cumulative recognition rate); Equal Error Rate (EER)
and the area below Receiver Operating Characteristic (ROCA) (smaller values mean
better results); first one recognition rate (First 1) that is achieved if only the first one
(most similar) image from the database is extracted (larger values mean better result). Percent (rank) 
of images that we need to extract from the database in order to achieve 100\% cumulative recognition rate
in the future we will denote as Cum100.
Graphical representation of the used characteristics is shown in Fig. ~\ref{fig_cmc} - ~\ref{fig_roc}.

\section{Experiments and results}

For recognition experiments we used the FERET database ~\citep{Phillips1998} containing greyscale
photographs of 1196 persons.
This database was collected in 1993-1996 at George Mason University
during the FERET (FacE REcognition Technology) program.
As far as we know, this is one of the largest databases of face photographs (of different persons)
in the world that is publicly available for face recognition research purposes ~\citep{Gross2005}.
This database is widely used for evaluating identification ~\citep{Phillips1998} and verification ~\citep{Rizvi1998}
performance of face recognition methods.

For recognition experiments we used 3541 facial images from this database.
The size of each image is 256x384 pixels, for each image this database contains manually selected eye
coordinates.
For training we used 1196 greyscale images from the $fa$ set of this database.
The same 1196 $fa$ images were used as a gallery (known persons),
and images from other sets (1195 $fb$ images, 722 $dup1$ images, 234 $dup2$ images, 194 $fc$ images)
were used as probes (unknown persons that we wish to recognize).
$fb$ set contain face images with different facial expressions,
$dup1$ and $dup2$ sets contain images that were taken
after some time interval (up to 1.5 years) from $fa$ images, and $fc$ set contains
images with different image input conditions (camera position and illumination).

At first we performed recognition experiments with $fa$ and $fb$ sets using
different number (100-1000) of PCA features
(different number of used eigenvectors, corresponding to the largest eigenvalues)
and compared the proposed masked Log-Gabor PCA with traditional
PCA (using cosine -based distance measure between "whitened" feature vectors).
Images were normalized using 3-point normalization method.
The results are presented in Table ~\ref{tbl_rec_my_comparison}.
The results showed, that first one recognition rate of masked Log-Gabor PCA (89.29-98.41\%)
is always higher than of traditional PCA (80.42-88.03\%), and
EER values of masked Log-Gabor PCA (0.33-1.59 \%) are always lower than EER values
of traditional PCA (1.92-4.26 \%).
Other characteristics (CMCA, ROCA, cumulative recognition) of masked Log-Gabor PCA are 
also better than of traditional PCA.
The results showed, that masked Log-Gabor PCA achieves larger first one recognition accuracy when we use larger
number of features (e.g., 100 PCA features - 89.29\% recognition accuracy, 1000 PCA features - 98.41\% recognition accuracy).
It must be noted, that masked Log-Gabor PCA uses shorter 
vectors (9240 Log-Gabor features) than traditional PCA (12646 greyscale features)
for PCA training.
Also we investigated another version of Log-Gabor PCA when Log-Gabor magnitude images
are not masked (only magnitudes, initial images are always masked).
In this case the sliding window search selects 10584 Log-Gabor features.
The results showed, that in some cases unmasked Log-Gabor PCA can achieve higher
first one recognition accuracy, but because these differences are not very large ($<$0.2\% with $>$200 features)
we prefer to use masked version of Log-Gabor PCA.

\begin{table*}[htbp]
\caption{Comparison of 3 face recognition methods: 1) Masked Log-Gabor PCA (MskLG), 2) UnMasked Log-Gabor PCA (UnMskLG), and 3) Traditional PCA (Trad).}
\label{tbl_rec_my_comparison}

\begin{tabular}{cccccccccccc}
\hline
Method & Feat.  &\multicolumn{6}{c}{Rank (\%) in order to achieve desired}	& CMCA  & First 1	& EER,	& ROCA  \\
       & num.   &\multicolumn{6}{c}{cumulative recognition accuracy}      	&       & rec., \%	& \%	&       \\
\cline{3-8}
		&		&	95	&	96	&	97	&	98	&	99	&	100				&		&		&			&		\\
\hline

MskLG	& 100	& 0.33	& 0.42	& 0.50	& 0.84	& 2.34	& 18.14	& 17.67	& 89.29	& 1.59	& 9.49	\\
UnMskLG	& 100	& 0.42	& 0.50	& 0.75	& 1.17	& 2.17	& 11.87	& 17.20	& 89.87	& 1.42	& 9.03	\\
Trad	& 100	& 0.84	& 1.17	& 1.59	& 2.93	& 6.44	& 36.87	& 37.27	& 83.85	& 2.26	& 28.77	\\
\hline

MskLG	& 200	& 0.17	& 0.25	& 0.33	& 0.42	& 0.84	& 6.69	& 11.34	& 93.56	& 0.75	& 2.48	\\
UnMskLG	& 200	& 0.17	& 0.17	& 0.25	& 0.42	& 0.75	& 6.94	& 11.44	& 93.56	& 0.75	& 2.61	\\
Trad	& 200	& 0.59	& 0.84	& 1.17	& 2.34	& 7.19	& 84.20	& 40.21	& 86.44	& 2.01	& 31.59	\\
\hline

MskLG	& 300	& 0.08	& 0.17	& 0.17	& 0.25	& 0.42	& 7.27	& 10.36	& 95.31	& 0.59	& 1.65	\\
UnMskLG	& 300	& 0.08	& 0.17	& 0.17	& 0.25	& 0.59	& 8.61	& 10.41	& 95.40	& 0.50	& 1.73	\\
Trad	& 300	& 0.59	& 0.84	& 1.34	& 2.84	& 5.52	& 93.65	& 39.36	& 87.87	& 1.92	& 29.91	\\
\hline

MskLG	& 400	& 0.08	& 0.08	& 0.17	& 0.17	& 0.33	& 3.76	& 9.46	& 96.99	& 0.42	& 0.93	\\
UnMskLG	& 400	& 0.08	& 0.08	& 0.08	& 0.17	& 0.33	& 6.44	& 9.58	& 97.15	& 0.42	& 1.04	\\
Trad	& 400	& 0.59	& 0.92	& 1.67	& 3.09	& 12.12	& 92.73	& 52.94	& 88.03	& 2.09	& 42.79	\\
\hline

MskLG	& 500	& 0.08	& 0.08	& 0.08	& 0.17	& 0.33	& 2.84	& 9.39	& 97.24	& 0.42	& 0.69	\\
UnMskLG	& 500	& 0.08	& 0.08	& 0.08	& 0.17	& 0.33	& 3.68	& 9.26	& 97.15	& 0.42	& 0.63	\\
Trad	& 500	& 0.92	& 1.25	& 2.01	& 4.10	& 16.64	& 74.50	& 58.73	& 87.53	& 2.34	& 48.26	\\
\hline

MskLG	& 600	& 0.08	& 0.08	& 0.08	& 0.17	& 0.25	& 2.26	& 9.07	& 97.74	& 0.33	& 0.48	\\
UnMskLG	& 600	& 0.08	& 0.08	& 0.08	& 0.17	& 0.25	& 1.42	& 8.91	& 97.74	& 0.33	& 0.39	\\
Trad	& 600	& 1.34	& 1.67	& 2.59	& 5.94	& 13.55	& 82.86	& 67.05	& 86.03	& 2.59	& 56.21	\\
\hline

MskLG	& 700	& 0.08	& 0.08	& 0.08	& 0.17	& 0.17	& 2.76	& 8.89	& 97.91	& 0.42	& 0.36	\\
UnMskLG	& 700	& 0.08	& 0.08	& 0.08	& 0.08	& 0.17	& 2.42	& 8.86	& 98.08	& 0.33	& 0.32	\\
Trad	& 700	& 1.42	& 2.26	& 4.35	& 8.86	& 22.41	& 95.07	& 88.53	& 85.02	& 3.18	& 77.03	\\
\hline

MskLG	& 800	& 0.08	& 0.08	& 0.08	& 0.17	& 0.17	& 3.51	& 8.98	& 97.99	& 0.33	& 0.38	\\
UnMskLG	& 800	& 0.08	& 0.08	& 0.08	& 0.08	& 0.17	& 3.18	& 9.02	& 98.08	& 0.33	& 0.38	\\
Trad	& 800	& 2.09	& 3.34	& 5.35	& 10.28	& 28.51	& 93.14	& 109.44& 83.09	& 3.51	& 97.12	\\
\hline

MskLG	& 900	& 0.08	& 0.08	& 0.08	& 0.08	& 0.25	& 1.84	& 8.92	& 98.16	& 0.33	& 0.33	\\
UnMskLG	& 900	& 0.08	& 0.08	& 0.08	& 0.17	& 0.17	& 1.76	& 8.85	& 98.08	& 0.33	& 0.28	\\
Trad	& 900	& 2.76	& 4.35	& 7.02	& 12.63	& 38.13	& 92.89	& 123.40& 82.43	& 3.93	& 110.99\\
\hline

MskLG	& 1000	& 0.08	& 0.08	& 0.08	& 0.08	& 0.17	& 6.35	& 9.44	& 98.41	& 0.33	& 0.63	\\
UnMskLG	& 1000	& 0.08	& 0.08	& 0.08	& 0.08	& 0.17	& 2.51	& 9.01	& 98.24	& 0.33	& 0.35	\\
Trad	& 1000	& 4.01	& 5.52	& 9.28	& 16.22	& 45.48	& 90.80	& 143.82& 80.42	& 4.26	& 130.80\\
\hline

\end{tabular}

\end{table*}

\begin{figure}[t]
\begin{minipage}[t]{0.5\linewidth}
\centering \includegraphics[height=76mm]{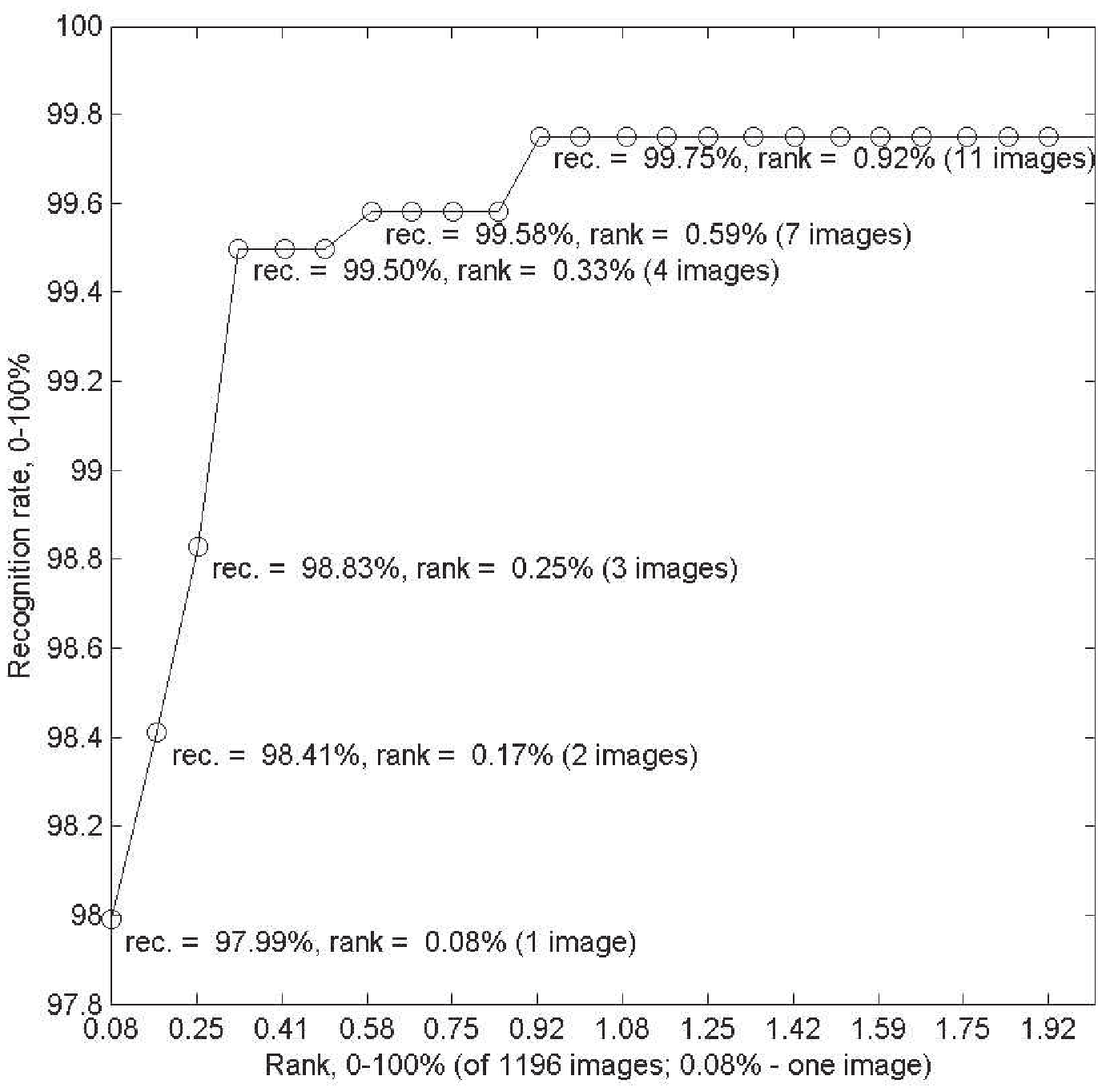}
\caption{CMC characteristic of Log-Gabor PCA method}
\label{fig_lgpca_cmc}
\end{minipage}%
\hspace{7mm}
\begin{minipage}[t]{0.5\linewidth}
\centering \includegraphics[height=76mm]{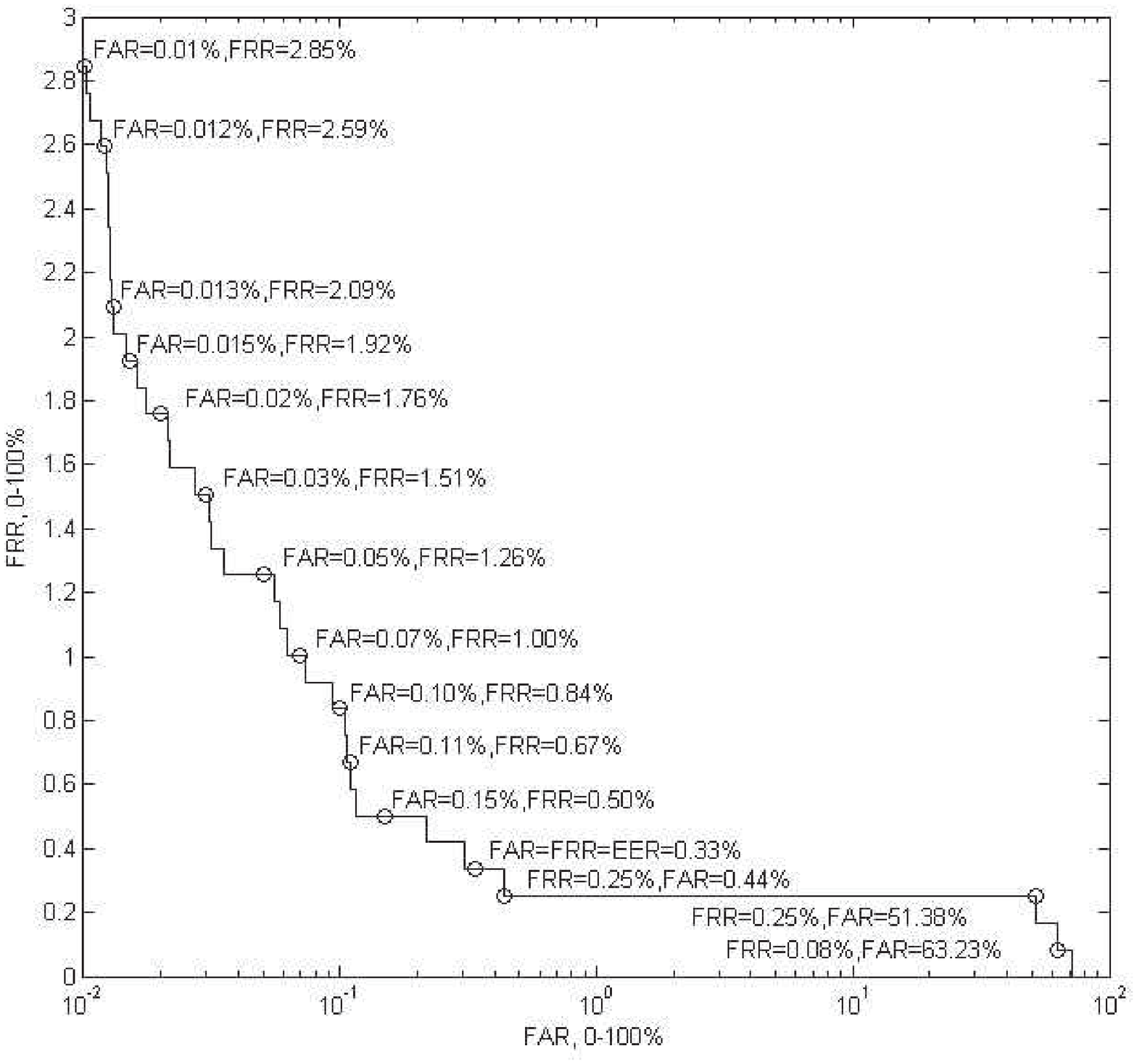}
\caption{ROC characteristic of Log-Gabor PCA method}
\label{fig_lgpca_roc}
\end{minipage}
\end{figure}

Also we compared our face recognition results with the results of other researchers.
For comparison we used the best results of the FERET program participants 
that took official FERET 1996-1997 tests ~\citep{Phillips2000}, ~\citep{NIST2001}.
Also we present the best results of some other researchers who tested their recognition methods
using all images (not subsets) from the FERET $fa$ (images of 1196 persons) and $fb$ (images of 1195 persons)
sets that contain faces with different facial expressions.
Most part of the compared methods for face normalization used manually located coordinates of eye centres.
These coordinates were marked by the creators of the FERET database and are distributed with this database. 
The results are summarized in Table ~\ref{tbl_rec_feret}.
Fully automatical methods are denoted by "auto", our different normalization methods are 
denoted by "3 pt." (3-point normalization) and "2 pt." (2-point normalization).
In Figures ~\ref{fig_lgpca_cmc} - ~\ref{fig_lgpca_roc} we also present CMC and ROC characteristics
of our Log-Gabor PCA method (900 PCA features, 2-point normalization).

\begin{sidewaystable}
\caption{Recognition of expression -variant faces from the FERET database (gallery contains 1196 $fa$ images, and probe contains 1195 $fb$ images).}
\label{tbl_rec_feret}
\begin{tabular}{lcccccccccc}
\hline
Method and its  &\multicolumn{6}{c}{Rank (\%) in order to achieve} & CMCA, & First 1 & EER, & ROCA, \\
authors  &\multicolumn{6}{c}{desired cumulative recognition, $(0,100\%]$}  & $[0,10^4]$ & recognition, & $[0,100\%]$ & $[0,10^4]$ \\
\cline{2-7}
						&	95	&	96	&	97	&	98	&	99	&	100		&			&	$[0,100\%]$	&			&			\\
\hline
MIT 1996 ~\citep{Moghaddam1996}	&	0.17&	0.25&	0.33&	1.09&	23.33&	99.83	&	84.14	&	94.81	&	4.77	&	203.25	\\
\hline
MIT 1996 auto~\citep{Phillips1998b}	&	-   &	-   &	-   &	-   &	-   &	-   	&	-   	&	$\sim$88.00	&	-	&	-	\\
\hline

UMD 1996 ~\citep{Etemad1997}	&	-	&	-	&	-	&	-	&	-	&	-		&	-		&	$\sim$83.50	&	$\sim$7.00	&	-		\\
\hline
UMD 1997 ~\citep{Zhao1998}	&	0.08&	0.08&	0.17&	0.33&	0.84&	75.92	&	18.91	&	96.23	&	1.09	&	14.37	\\
\hline
USC 1997 ~\citep{Okada1998}&	0.17&	0.17&	0.25&	0.33&	3.09&	50.67	&	27.94	&	94.98	&	2.51	&	57.52	\\
\hline
USC 1997 auto ~\citep{Okada1998}&	- &	- &	- &	- &	- &	-	&	-	&	94.00	&	-	&	-	\\
\hline

MSU 1996 ~\citep{Swets1996}	&	-	&	-	&	-	&	-	&	-	&	-		&	-		&	$\sim$88.50	&	$\sim$3.00	&	-		\\
\hline
Bayesian MAP ~\citep{Teixeira2003} &	1.92&	2.51&	4.18&	6.52&	13.8&	70.15	&	67.11	&	81.92	&	-		&	-		\\
\hline
EBGM Standard ~\citep{Bolme2003} &	0.59&	0.92&	1.34&	2.42&	9.2	&	37.54	&	34.26	&	88.37	&	-		&	-		\\
\hline
EBGM Optimised ~\citep{Bolme2003} &	-	&	-	&	-	&	-	&	-	&	-	&	-	&	89.80	&	-		&	-		\\
\hline
PCA MahCosine ~\citep{CSU2003} &	0.84&	1.17&	2.26&	4.43&	10.28&	60.45	&	48.90	&	85.27	&	-		&	-		\\
\hline
Gabor features ~\citep{Kepenekci2002} &	-	&	-	&	-	&	-	&	-	&	-		&	-		&	96.30	&	-		&	-		\\
\hline

Haar+AdaBoost ~\citep{Jones2003} &	-	&	-	&	-	&	$\sim$0.42	&	$\sim$1.17	&	-		&	-		&	$\sim$94.00	&	$\sim$1.00		&	-		\\
\hline
Gabor+AdaBoost ~\citep{Yang2004} &	-	&	-	&	-	&	-	&	-	&	-		&	-		&	$\sim$95.20	&	-		&	-		\\
\hline
SOM ~\citep{Tan2005} &	-	&	-	&	-	&	-	&	-	&	-		&	-		&	$\sim$91.00	&	-		&	-		\\
\hline

Our Log-Gabor PCA, 900 PCA feat., 3 pt.	& 0.08 & 0.08 & 0.08 & 0.08 & 0.25 & 1.84  & 8.92  & 98.16 & 0.33 & 0.33 \\
\hline
Our Log-Gabor PCA, 900 PCA feat., 2 pt.	& 0.08 & 0.08 & 0.08 & 0.17 & 0.33 & 68.81 & 24.02 & 97.99 & 0.33 & 15.78 \\
\hline

Our trad. PCA, 900 PCA feat., 3 pt.	& 2.76	& 4.35	& 7.02	& 12.63	& 38.13	& 92.89	& 123.40 & 82.43 & 3.93	& 110.99 \\
\hline
Our trad. PCA, 900 PCA feat., 2 pt.	& 6.94  & 8.61  & 13.21 & 16.64 & 31.44 & 99.58 & 149.75 & 76.90 & 5.27	& 136.24 \\
\hline

Our Log-Gabor PCA 4x4, 900 PCA feat., 3 pt.	& 0.08 & 0.08 & 0.08 & 0.08 & 0.08 & 0.17 &	0.59 &	98.49	&	0.17	&	0.14		\\
\hline

\end{tabular}
\end{sidewaystable}

Now we will briefly describe face recognition methods of other researchers
and will compare achieved results.
MIT 1996 (Massachusetts Institute of Technology, MIT Media Laboratory) method was developed
by ~\citep{Moghaddam1996}. For recognition they used dual (intrapersonal and extrapersonal)
PCA and Bayesian MAP (maximum a posteriori) similarity measure. For learning were used image
pairs of the same and different persons.
UMD 1996, UMD 1997 (University of Maryland) face recognition methods are based on 
PCA and LDA (Linear Discriminant Analysis) and were developed by
~\citep{Etemad1997, Zhao1998}.
For training were used several images per person, for recognition were used 300 features. 
USC 1997 (University of Southern California) method was developed by
~\citep{Wiskott1997, Okada1998}.
For recognition they used Gabor Jets and Elastic Bunch Graph Maching (EBGM).
Faces were resized to 128x128 pixels and normalized using histogram equalization
For recognition were used about 1920 features that correspond to 40 Gabor filters
(5 scales and 8 orientations) at 48 graph nodes.
MSU 1996 (Michigan State University) method was developed by ~\citet{Swets1996}.
For recognition they used PCA and LDA.
~\citet{CSU2003} Bayesian MAP ~\citep{Teixeira2003}, EBGM Standard, and EBGM Optimised ~\citep{Bolme2003}
face recognition methods were developed by the researchers at Colorado State University.
These methods are similar to the corresponding methods developed at
MIT and USC. The ~\citet{CSU2003} PCA MahCosine method is a traditional PCA with cosine -based distance measure between "whitened"
feature vectors.
~\citet{CSU2003} for recognition used 130x150 images, faces were masked using elliptical mask, unmasked
image part was normalized using histogram equalization. CSU EBGM method for recognition extracts 
more than 6000 features (80 Gabor features x 80 graph points).
~\citet{Kepenekci2002} for recognition used magnitudes of Gabor filters and
similar distance measures as were used by ~\citep{Wiskott1997}. For recognition were extracted 40
Gabor features (5 scales and 8 orientations) and 2 coordinates of these features at varying number of face points.
~\citet{Jones2003} used Haar -like features and AdaBoost training. For recognition were used
45x36 images without masking. The use small images may be related with the fact that
AdaBoost training requires huge computational resources.
~\citet{Yang2004} used Gabor features and AdaBoost training -based recognition method. 
~\citet{Tan2005} for face recognition used Self-Organizing Map (SOM) and
soft k nearest neighbor (soft k-NN) ensemble method.

As we can see from the Table ~\ref{tbl_rec_feret}, the highest first one recognition accuracy was
achieved by the following methods:
our Log-Gabor PCA (98.16\% using 2-point normalization and  98.49\% using 3-point normalization),
Gabor features ~\citep{Kepenekci2002} -based method (96.30\%), and UMD 1997 ~\citep{Zhao1998} PCA+LDA -based method (96.23\%).
The best EER results were achieved by our Log-Gabor PCA (0.33\%), 
Haar+AdaBoost ~\citep{Jones2003} method ($\sim$1.00), and UMD 1997 ~\citep{Zhao1998} PCA+LDA -based method (1.09\%).  
It is interesting to note that our traditional PCA with 2-point normalization achieves
lower recognition accuracy than traditional PCA of ~\citep{CSU2003}.
But when we combine our traditional PCA with Log-Gabor features,
our method achieves higher recognition accuracy that many other methods.
Also we can note that PCA with grayscale features is much more sensitive to the chosen
image normalization method (76.90\% first one recognition using 2-point normalization and 82.43\% using 3-point normalization)
than our Log-Gabor PCA (98.16\% using 2-point normalization and  98.49\% using 3-point normalization).
Using 3-point normalization faces are masked and cropped more accurately
than using 2-point normalization, and recognition results using Log-Gabor PCA and
3-point normalization are also better. These differences are especially visible
when we compare Cum100, CMCA and ROCA values of 2-point ant 3-point methods.
It is interesting to note, that EBGM -based methods perform positioninig of graph
nodes around the face also enough accurately and this may be one of the reasons why
EBGM -based methods and Log-Gabor PCA method using 3-point normalization achieve
better Cum100 results than other methods that use 2-point normalization.
Using our method with 3-point normalization in order to achieve 100\% cumulative recognition rate
we need to extract from the database only 1.84\% of images (that is 1196*1.84/100 = 22 images), and
using CSU EBGM method we need to extract 37.54\% of images (that is 449 images).
In the last line of the
Table ~\ref{tbl_rec_feret} we present the results of our method if we use 3-point normalization,
4x4 sliding window (without overlapping), masking,
and 19704 Log-Gabor magnitude features (the number of PCA features remains 900).
As we can see from these  results, using larger number of Log-Gabor features
(smaller sliding window) we can achieve even better cumulative recognition
results than using 8x8 window with 2 pixels overlapping. That is in order to achieve 
99\% cumulative recognition rate we need to extract from the database only 2 images (0.17\%),
and in order to achieve 100\% cumulative recognition rate we need to extract 7 images (0.59\%).

Because in real life situations face recognition methods are usually used with automatically detected
faces and facial features. So it is desirable to know how detection errors affect
recognition accuracy, what recognitiom method is less sensitive to feature detection errors.
Using this information we can decide how accurately faces and facial features should be detected
in order to achieve desirable recognition accuracy.
In the Table ~\ref{tbl_rec_feret} we presented some results of other researchers that
used automatical detection of faces and facial features (notation "auto").
USC 1997 automatic method ~\citep{Okada1998} for loacation of face and facial features
(eyes, nose, lips, face contour) used EBGM with small number of graph nodes (16 nodes).
As it is stated by the authors, their method locates facial features very accurately, so
the difference between recognition results using manually and automatically detected features
is $\sim$1\%.
MIT 1996 ~\citep{Moghaddam1996} automatical method for detection of eyes and lips used PCA-based detector
and probabilistic verification of detected feature locations. Automatical method
achieved $\sim$7\% lower recognition accuracy than the same method that used manual feature detection.
The results showed that fully automatical USC 1997 method can achieve
much higher recognition accuracy than MIT 1996 method.
But because the authors used different methods for detecting faces and facial features and did
not present any quantitative information about feature detection accuracy, 
we cannot say for sure what recognition method (that was tested without automatical detection)
it is better to use with automatical feature detection method.
It is possible that one recognition method is less sensitive to feature detection errors
than another, but also it is possible that the main differences are only in feature detection methods
and their feature detection accuracy.

In order to find out how sensitive is our face recognitiom method to feature detection errors
and not to bind to concrete feature detection method we manually shifted the markers of eye centres
using pre-defined shift directions and distances.
For this experiment we used all facial images from the FERET $fa$ and $fb$ sets.
We shifted only eye markers of $fb$ images using 4 shift directions (0, $\pi/2$, $\pi$, $3\pi/2$) and
the following shift distances: 0\%, 2\%, 4\%, 6\%, 8\%, 10\%, 12\%.
Recognition accuracy using each shift distance was calculated as an average of 4 results that
correspond to 4 directions.
Shift distance is calculated as a percentage of the distance between manually selected eye centres.
For example, if the distance between manually selected eyes is 100 pixels and we
wish to use 4\% shift, then these 4\% will correspond to 4 pixels. The results that
were achieved using 0\% shift (it means that no shift is performed and we simply use
manually selected eye coordinates) were used as a baseline for comparison with the results that were
achieved using other shift distances.
The results are presented in Table ~\ref{tbl_rec_sensivity}, where notations
First1d and EERd mean absolute differences between achieved recognition result using some shift
and the result without any shift (baseline). For experiments was used 2-point normalization method.

\begin{table*}[htbp]
\caption{Face recognition accuracy using shifted markers of eye centres (simulated eye location errors).}
\label{tbl_rec_sensivity}
\begin{tabular}{llcccccccc}
\hline
Shift & Method & \multicolumn{4}{c}{Shifted left eye marker} & \multicolumn{4}{c}{Shifted right eye marker} \\
\cline{3-6} \cline{7-10} 
size, \% & name & First1 & First1d & EER & EERd & First1 & First1d & EER & EERd \\
\hline
0\% & PCA & 76.90 & - & 5.27 & - & 76.90 & - & 5.27 & - \\
    & Log-Gabor PCA & 97.99 & - & 0.33 & - & 97.99 & - & 0.33 & - \\
\hline
2\% & PCA & 73.10 & 3.80 & 6.05 & 0.78 & 72.95 & 3.95 & 6.05  & 0.78 \\
    & Log-Gabor PCA & 97.45 & 0.54 & 0.40 & 0.07 & 97.41 & 0.58 & 0.40 & 0.07 \\
\hline
4\% & PCA & 62.45 & 14.45 & 8.39 & 3.12 & 61.86 & 15.04 & 8.56 & 3.29 \\
    & Log-Gabor PCA & 96.76 & 1.23 & 0.44 & 0.11 & 96.88 & 1.11 & 0.50 & 0.17 \\
\hline
6\% & PCA & 47.78 & 29.12 & 12.32 & 7.05 & 45.04 & 31.86 & 12.59 & 7.32 \\
    & Log-Gabor PCA & 95.65 & 2.34 & 0.65 & 0.32 & 95.31 & 2.68 & 0.63 & 0.30 \\
\hline
8\% & PCA & 31.17 & 45.73 & 16.38 & 11.11 & 31.30 & 45.60 & 17.22 & 11.95 \\
    & Log-Gabor PCA & 93.41 & 4.58 & 0.96 & 0.63 & 92.95 & 5.04 & 0.96 & 0.63 \\
\hline
10\% & PCA & 19.21 & 57.69 & 21.51 & 16.24 & 19.29 & 57.61 & 22.53 & 17.26 \\
     & Log-Gabor PCA & 90.02 & 7.97 & 1.46 & 1.13 & 89.02 & 8.97 & 1.55 & 1.22 \\
\hline
12\% & PCA & 11.40 & 65.50 & 26.46 & 21.19 & 11.92 & 64.98 & 27.13 & 21.86 \\
     & Log-Gabor PCA & 84.29 & 13.70 & 2.24 & 1.91 & 83.24 & 14.75 & 2.36 & 2.03 \\
\hline

\end{tabular}

\end{table*}

The results (Table ~\ref{tbl_rec_sensivity}) showed that our Log-Gabor PCA
is less sensitive to feature detection errors than
traditional PCA and can achieve 89-90\% recognition accuracy even if one eye
is shifted by 10\%. In order to create fully automatical face recognition method and
achieve similar recognition accuracy that
was achieved by USC 1997 ~\citep{Okada1998} fully automatical method
we should combine our recognition method
with automatical eye detection method that detects the centres of eyes with smaller than 6\% shifts
(total shift for both images) when compared to manually selected eye centres.
Those readers who are interested in automatical face and facial features detection methods 
can find an overview of such methods in ~\citep{Yang2002} and
~\citep{Perlibakas2003}.

In real life situations the accuracy of face recognition is also affected by many other factors
like aging and manual change of appearance (hairstyle, makeup), image input (camera position)
and illumination conditions. It is natural that we cannot have the same looking faces and the 
same imaging conditions after a longer time period. 
In order to find out how recognition accuracy is affected by these factors we 
performed recognition experiments using the following probe sets of images from the FERET database:
$dup1$ - 722 images of 243 persons, at least
2 images with different expressions per person, time interval from $fa$ images is 0-34 months,
photographs of 166 persons were taken after some period of time (not the same day than $fa$ images);
$dup2$ - 234  images of 75 persons, at least 2 images with different expressions per person,
time interval from $fa$ images is more than 18 months;
and $fc$ - 194  images of 194 persons that were acquired on the same day, but with different camera
position and illumination.

\begin{table*}[htbp]
\caption{The influence of aging and illumination to face recognition accuracy using 1196 $fa$ gallery images (FERET database) and the following probe sets: $dup1$, $dup2$, $fc$.}
\label{tbl_rec_feret_dup1_dup2_fc}
\begin{tabular}{lcccccc}
\hline
Method and its authors & \multicolumn{3}{c}{First one rec., $[0,100\%]$	} & \multicolumn{3}{c}{ EER, $[0,100\%]$	} \\
\cline{2-7}
                       & dup1 &  dup2 & fc & dup1 & dup2 & fc  \\
\hline

MIT 1996   ~\citep{NIST2001}	&	57.60   &	34.20   &	32.00   &	17.70   &	21.20   &	18.00	\\
\hline
MIT 1996 auto ~\citep{Phillips1998b}	&	$\sim$50.00   &	-   &	-   &	-   &	-   &	-  \\
\hline
UMD 1996 ~\citep{Phillips1998b} &	$\sim$32.00	&	$\sim$9.00	&	$\sim$30.00	&	-	&	-	&	-\\
\hline
UMD 1997 ~\citep{NIST2001}	&	47.20	&	20.90	&	58.80	&	12.60	&	13.40	&	10.00\\
\hline
MSU 1996 ~\citep{Phillips1998b}	&	$\sim$33.00	&	$\sim$17.00		&	$\sim$32.00		&	-	&	-	&	- \\
\hline
USC 1997 ~\citep{NIST2001} &	59.10 &	52.10 &	82.00 &	13.30 &	14.20 &	5.10	\\
\hline
USC 1997 ~\citep{Okada1998}&	62.00 &	52.00 &	82.00 &	- &	- &	-	\\
\hline
USC 1997 auto ~\citep{Okada1998}&	61.00 &	52.00 &	80.00 &	- &	- &	- \\
\hline
Gabor feat. ~\citep{Kepenekci2002} &	58.30	&	47.40	&	69.60	&	-	&	-	&	- \\
\hline

Our trad. PCA, 900 PCA feat., 2 pt.	& 44.74 & 35.04 & 62.89 & 13.99 & 19.03 & 9.79 \\
\hline
Our Log-Gabor PCA, 900 PCA feat., 2 pt.	& 72.44 & 65.81 & 90.21 & 3.60 & 4.70 & 1.03	\\
\hline

\end{tabular}

\end{table*}

The results (Table ~\ref{tbl_rec_feret_dup1_dup2_fc}) showed that our Log-Gabor PCA method achieves
8-10\% higher recognition accuracy and at least 4\% lower EER than other compared methods.
Our recognition results
showed that even using single training image per person we can improve recognition accuracy
of face images that were took after longer time period.
But also it is obvious that different imaging conditions after longer time period significanly
reduce face recognition accuracy of all compared methods and that for such difficult tasks
we need better image normalization and feature extraction techniques.

We also performed several face recognition experiments using the AR database ~\citep{Martinez1998}.
This database was created by A. Martinez and R. Benavente at Computer Vision Center, Purdue University in 1998.
It contains facial photographs of 126 persons with strictly controlled facial expressions and lighting.
The size of images is 768x576 pixels.
Images of each person were captured in two sessions (s1, s2) that were separated by two weeks time.
From this database we used 1610 images of 115 persons (14 images per person = 2 sessions x 7 images
per session).
We used the following images: neutral (ne), happy (ha), angry (an), and screaming (sc) expressions;
neutral expression with left illumination source (lis) turned on, right illumination source (ris) turned on, and both illumination sources (bis) turned on.  
For training and as a galery set we used 115 images with neutral expression from the first session (s1ne).
For recognition we used the following probe sets that correspond to different type of transformation
(neutral, expression, illumination) ~\citep{Wang2003}: s1expr (s1ha, s1an, s1sc images),
s1illum (s1lis, s1ris, s1bis images),
s2neutral (s2ne images), s2expr (s2ha, s2an, s2sc images), and s2illum (s2lis, s2ris, s2bis images).
First one recognition results using these sets are presented in Table ~\ref{tbl_rec_ar}.
The last lines of this table also present EER results of our methods.
For experiments was used 2-point normalization method.

\begin{table*}[htbp]
\caption{Face recognition results using AR database.}
\label{tbl_rec_ar}
\begin{tabular}{lcccccc}
\hline
Method and  its authors &  \multicolumn{5}{c}{Firs one rec. results using different probe sets} \\
\cline{2-6}
                        & s1expr & s1illum & s2neutral & s2expr & s2illum \\

\hline
PCA ~\citep{Wang2003} & - & - & 84.4 & 56.7 & 24.4	\\
\hline
EBGM Gabor features ~\citep{Wang2003}	& - & - & 86.7 & 66.7 & 52.2	\\
\hline
EBGM Gabor features	+ &  &  &  &  & 	\\
Bayes matching ~\citep{Wang2003} & - & - & 93.3 & 86.0 & 86.7	\\
\hline
PCA ~\citep{Martinez2003a}	& 72.00 & - & - & - & - \\
\hline
Correlation ~\citep{Martinez2003a}	& 74.33 & - & - & - & - \\
\hline
PCA + optical flow ~\citep{Martinez2003a}	& 83.00 & - & - & - & - \\
\hline
Motion estimation ~\citep{Martinez2003b}	& 84.67 & - & - & - & - \\
\hline
Our traditional PCA, 100 feat., 2 pt.	& 70.43 & 62.90 & 92.17 & 58.52 & 46.67	\\
\hline
Our Log-Gabor PCA, 100 PCA feat., 2 pt.	& 85.51 & 82.90 & 99.13 & 77.39 & 63.48	\\
\hline
 & \multicolumn{5}{c}{EER results of our methods} \\
\hline
Our traditional PCA, 100 feat., 2 pt.	& 6.96 & 5.51 & 2.61 & 11.02 & 10.72	\\
\hline
Our Log-Gabor PCA, 100 PCA feat., 2 pt.	& 3.48 & 3.19 & 0.87 & 6.67 & 6.67	\\
\hline

\end{tabular}

\end{table*}

The results (Table ~\ref{tbl_rec_ar}) showed that our method achieves slightly higher recognition accuracy than other compared methods
that used single training image (with neutral expression) per person.
It is important to note that the comparison of different methods in the Table ~\ref{tbl_rec_ar}
is not very exact, because for experiments different authors used different number of images:
~\citep{Wang2003} used images of 90 persons, ~\citep{Martinez2003a}, ~\citep{Martinez2003b} used images of 100 persons, 
and we used images of 115 persons.
The results showed that EBGM Gabor features (features are extracted at graph nodes) and Bayes matching -based algorithm
~\citep{Wang2003} can achieve much higher recognition accuracy than our method and all
other methods (our method achieved better results only when recognizing faces with neutral expressions).
But for training of this method we need multiple images per person,
and this method was trained using 7 images per person from the first session
with different expressions and illumination conditions.
All other compared methods for training used single image per person (image with neutral expression
from the first session).
It is interesting to note PCA + optical flow ~\citep{Martinez2003a} and Motion estimation ~\citep{Martinez2003b} -based recognition methods
were specially designed for recognizing expression -variant faces, and these methods use weigting of facial features
in order to reduce the influence of changed expression to the accuracy of face recognition.
Perhaps these weigting methods could improve recognition accuracy of our face recognition
method, and in the future we are going to investigate different feature weighting and masking
methods in order to improve recognition accuracy of expression -variant faces.   

\section{Conclusions and future work}

In this article we proposed a novel face recognition method based on Principal Component Analysis (PCA)
and Log-Gabor filters.
The experiments showed that using the proposed combination of Log-Gabor features and
sliding window -based feature selection method,
Principal Component Analysis, "whitening", and cosine -based distance measure we can achieve very
high recognition accuracy
(97-98\%) and low error rates (0.3-0.4\% Equal Error Rate)
using the FERET database that contains photographs of more than 1000 persons.
The results of our algorithm are among the best results that were ever achieved using this database.
In the future we are going to investigate the possibilities of using decomposed
Log-Gabor feature vectors
and multiple PCA spaces in order to have the possibility of using this method with an
unlimited number of training images.
Because the results of all compared methods showed that the accuracy of face recognition is very
affected by the lighting conditions, in the future we are going to investigate different lighting
normalization methods and test them with the Log-Gabor PCA face recognition method.

\section{Acknowledgements}

Portions of the research in this paper use the FERET database of facial images collected under the FERET program.


\end{document}